\documentclass[10pt,twocolumn,letterpaper]{article}  % TODO: for pdf compiler
\usepackage{cvpr}              % To produce the CAMERA-READY version
\newcommand{\thirdheading}[1]{
  \vspace{6pt}
  \noindent{\textbf{#1.}}~
}

\usepackage{color} % Additional
\usepackage{amssymb} % Additional
\usepackage{multirow} % Additional
\definecolor{cvprblue}{rgb}{0.21,0.49,0.74}
\usepackage[pagebackref,breaklinks,colorlinks,allcolors=cvprblue]{hyperref}

\def\paperID{} % *** Enter the Paper ID here
\def\confName{CVPR}
\def\confYear{2026}

\title{Beyond Pedestrians: Caption-Guided CLIP Framework \\
for High-Difficulty Video-based Person Re-Identification}

\author{Shogo Hamano, Shunya Wakasugi, Tatsuhito Sato, Sayaka Nakamura\\
Sony Group Corporation\\
{\tt\small \{shogo.hamano, shunya.wakasugi, tatsuhito.sato, sayaka.nakamura\}@sony.com}
}

\begin{document}
\maketitle
\begin{abstract}

In recent years, video-based person Re-Identification (ReID) has gained attention for its ability to leverage spatiotemporal cues to match individuals across non-overlapping cameras.
However, current methods struggle with high-difficulty scenarios, such as sports and dance performances, where multiple individuals wear similar clothing while performing dynamic movements.
To overcome these challenges, we propose CG-CLIP, a novel caption-guided CLIP framework that leverages explicit textual descriptions and learnable tokens.
Our method introduces two key components: Caption-guided Memory Refinement (CMR) and Token-based Feature Extraction (TFE).
CMR utilizes captions generated by Multi-modal Large Language Models (MLLMs) to refine identity-specific features, capturing fine-grained details.
TFE employs a cross-attention mechanism with fixed-length learnable tokens to efficiently aggregate spatiotemporal features, reducing computational overhead.
We evaluate our approach on two standard datasets (MARS and iLIDS-VID) and two newly constructed high-difficulty datasets (SportsVReID and DanceVReID).
Experimental results demonstrate that our method outperforms current state-of-the-art approaches, achieving significant improvements across all benchmarks.

\end{abstract}    
\section{Introduction}
\label{sec:intro}

Video-based person Re-Identification (ReID)~\cite{liu2015spatio, liu2021grl, zhang2020mgrafa}, which aims to identify individuals across video sequences, has attracted significant attention over the past decade.
Numerous methods have been developed to address this task, including approaches based on CNNs~\cite{chen2018snippet, dai2018tempresi, fu2019sta, yang2024stfe} and Transformers~\cite{alsehaim2022vidtrans, liu2024tmt, tang2022mstat, zang2022pit}.
Recent works~\cite{li2023clipreid, zhang2024vsla, yu2024tf, li2023pcl} have explored leveraging large-scale pre-trained Vision-Language Models (VLMs), such as Contrastive Language-Image Pre-training (CLIP)~\cite{radford2021clip}, to enhance performance.
For example, CLIP-ReID~\cite{li2023clipreid} applies CLIP to image-based person ReID by optimizing text tokens through prompt learning.
TF-CLIP~\cite{yu2024tf} introduces a one-stage text-free approach for video-based person ReID, where identity-specific image features (CLIP-Memory) are computed from all training data and used as a substitute for text features.

% h: here, t: top, b: bottom, p: independent page
\begin{figure}[tbp]
    \centering
    \includegraphics[width=0.9\linewidth]{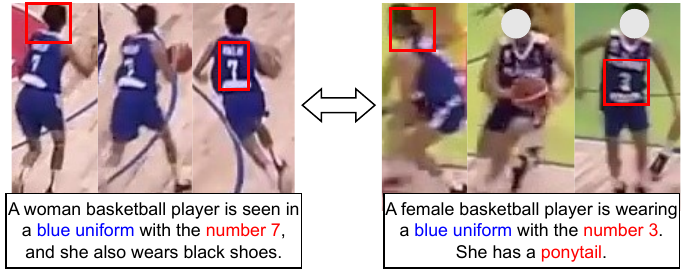}
    \caption{
    Two basketball players with similar appearances.
    They wear identical uniforms and shoes of similar colors but differ in their hairstyles and jersey numbers.
    These subtle differences can be explicitly represented in textual descriptions.
    }
    \label{fig:dataset}
\end{figure}

\begin{figure}[tbp]
    \centering
    \includegraphics[width=1.0\linewidth]{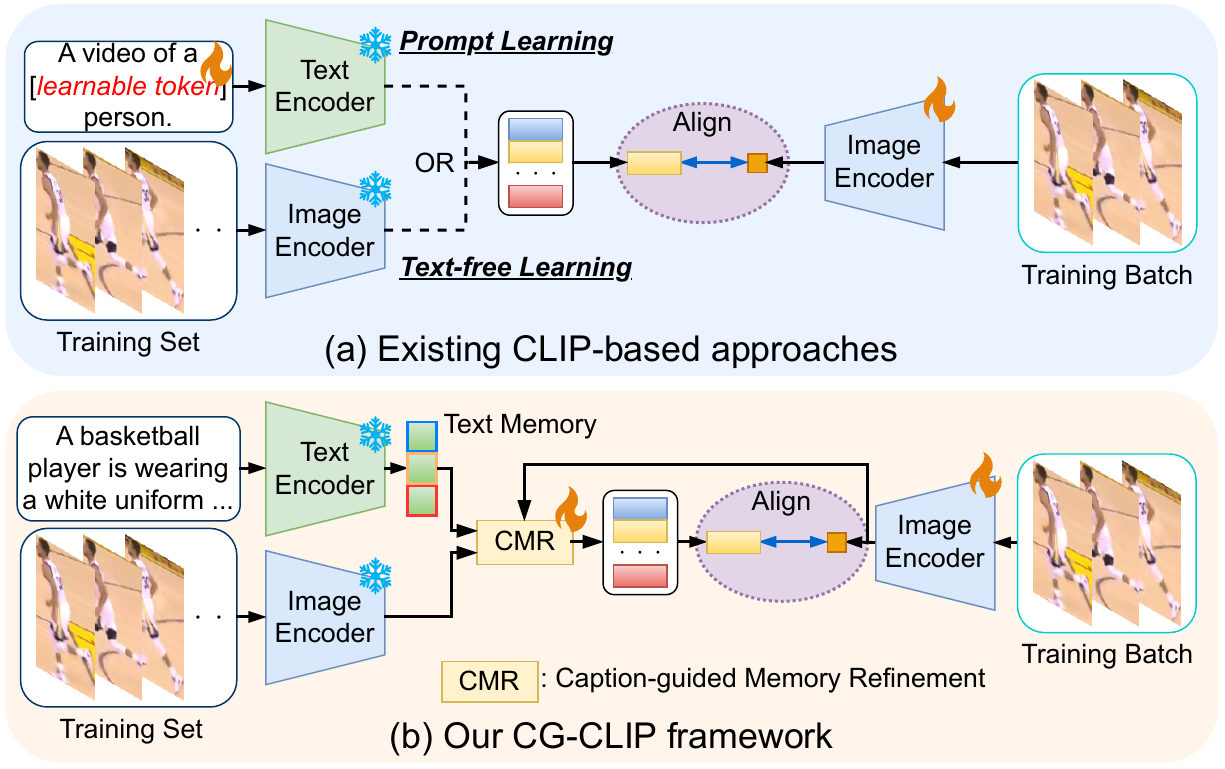}
    \caption{Comparison of existing approaches and our method.
    (a) Prompt learning and text-free learning approaches.
    (b) Our proposed CG-CLIP framework that leverages text captions.}
    \label{fig:method_overview}
\end{figure}

Despite these advancements, most existing methods focus on relatively straightforward scenarios, such as videos of pedestrians captured by surveillance cameras.
In contrast, real-world applications of person ReID, such as tracking~\cite{de2024contrastr, zhang2021fairmot}---where person ReID is crucial for long-term identity association---often involve more complex scenarios including sports or dance performances~\cite{huang2024deepieou, sun2024gta}, where individuals are dressed in similar uniforms or costumes and move rapidly. 
When faced with such scenarios, current methods often struggle due to two main challenges: (i) capturing fine-grained discriminative features and (ii) efficiently aggregating spatiotemporal information.

\Cref{fig:dataset} depicts an example of two visually similar individuals, along with textual descriptions highlighting their unique characteristics.
At first glance, it is challenging to distinguish them; however, close inspection reveals subtle differences in hairstyles and jersey numbers.
By representing these attributes in textual form, nuanced differences that are often overlooked become more prominent and easier to identify.
This observation underscores the potential of textual information as a complementary resource for enhancing person ReID performance in high-difficulty domains.
Despite this, most existing methods do not explicitly utilize such descriptions and instead opt for using learnable tokens or image features, as shown in \cref{fig:method_overview} (a).
These approaches make it difficult to capture fine-grained distinctions that appear in only small regions of the image.

The effective exploitation of robust spatiotemporal features also plays a crucial role in video-based person ReID. 
Previous methods~\cite{yu2024tf, yang2024stfe, wang2024top} employ self-attention mechanisms to aggregate information across temporal and spatial dimensions.
However, this approach has a drawback in terms of computational cost, which scales quadratically with the input token length.
This issue becomes particularly evident in dynamic scenes, where increased motion often requires higher frame rates or higher-resolution images, substantially increasing memory and computational costs.

To address these issues, we propose CG-CLIP, a novel caption-guided CLIP framework that leverages explicit textual descriptions and learnable tokens to capture robust representations.
Our method consists of two key components: Caption-guided Memory Refinement (CMR) and Token-based Feature Extraction (TFE).
We first synthesize training captions that describe subtle attributes (\eg, hairstyle, jersey number, and shoes) using Multi-modal Large Language Models (MLLMs)~\cite{abouelenin2025phi4}.
Next, as illustrated in \cref{fig:method_overview} (b), we encode all captions with a frozen CLIP text encoder and build an identity-specific Text Memory (\ie, one text prototype per identity) by averaging caption embeddings.
Unlike instance-wise image--text alignment, CMR uses the Text Memory as queries to selectively attend to identity-discriminative patch tokens from the training batch.
It then refines the contrastive memory target by injecting caption-guided fine-grained cues into the image-based CLIP-Memory.
Captions are not required during inference, which maintains the efficiency of the proposed method.

Furthermore, we propose the TFE module for effective extraction of spatiotemporal information.
In the TFE module, a cross-attention mechanism with learnable tokens is applied sequentially along the temporal and spatial dimensions to all tokens output by the image encoder.
This design keeps the attention cost linear in the number of input tokens, making it more scalable when denser temporal sampling or higher-resolution inputs are needed in fast-motion scenarios.
We evaluate our approach on two standard benchmarks, MARS and iLIDS-VID.
Additionally, we propose two high-difficulty datasets, named SportsVReID and DanceVReID.
These datasets are generated from Multi-Object Tracking (MOT) datasets (SportsMOT~\cite{cui2023sportsmot} and DanceTrack~\cite{sun2022dancetrack}), containing scenes of sports and dance performances.
Experiments on those datasets demonstrate the performance improvements achieved by our method.

In summary, our main contributions are as follows:
\begin{itemize}
    \item We propose CG-CLIP, a novel caption-guided CLIP framework with Caption-guided Memory Refinement (CMR), which is capable of extracting fine-grained features using explicit captions.
    \item We develop a Token-based Feature Extraction (TFE) module to integrate spatiotemporal features while keeping computational overhead low.
    \item Extensive experiments demonstrate that our method consistently outperforms state-of-the-art approaches on both existing benchmarks and our newly constructed datasets.
\end{itemize}

\section{Related work}
\label{sec:related_works}

\subsection{Person ReID datasets}
The person ReID task can be categorized into image-based person ReID~\cite{he2021transreid, yang2024promptsg, zhai2024mpreid, zhou2019osnet} and video-based person ReID~\cite{chen2018snippet, tang2022mstat, zhang2024vsla, zheng2016mars}.
In both categories, most existing datasets focus on pedestrian scenarios.
For image-based person ReID, popular datasets include Market-1501~\cite{zheng2015market}, MSMT17~\cite{wei2018msmt}, and DukeMTMC~\cite{zheng2017dukemtmcreid}.
In recent years, datasets containing sports scenes have also been proposed, such as the Player Re-Identification dataset~\cite{van2022playerreid} and SoccerNet~\cite{Giancola2022soccernet}, which consist of basketball and soccer scenes, respectively.
% For instance, the Player Re-Identification dataset, collected from basketball game videos, consists of 9,529 images of 486 identities.
In the case of video-based person ReID, widely used benchmarks include PRID~\cite{hirzer2011prid}, iLIDS-VID~\cite{wang2014ilids}, MARS~\cite{zheng2016mars}, and LS-VID~\cite{li2019lsvid}.
In contrast to image-based person ReID, there are currently few datasets comprising scenes such as sports or dance performances that contain multiple individuals with similar appearances.
To address this gap, we introduce two new video-based person ReID datasets that specifically target such high-difficulty settings.

\subsection{Video-based person ReID methods}
Video-based person ReID methods aim to utilize both spatial and temporal cues to extract discriminative features from videos of persons.
Early studies have explored various approaches, including methods based on RNNs/LSTMs~\cite{mclaughlin2016recurrent, dai2018tempresi}, 3D convolutions~\cite{gu2020appearance, li2019multiscale3d, aich2021strf}, temporal pooling~\cite{chen2018snippet, fu2019sta, zheng2016mars}, and attention mechanisms~\cite{liu2021grl, liu2023lstrl, chen2022sgmn}.  
For example, Chen \etal~\cite{chen2022sgmn} propose a region-level saliency and granularity mining network to extract temporal invariant features.
In our method, we propose a TFE module that extracts robust temporal information through a cross-attention mechanism with learnable tokens.

In recent years, Transformer-based approaches~\cite{liu2024tmt, he2021dil, wu2024tcvit, zang2022pit} have outperformed previous CNN-based methods.
Zang \etal~\cite{zang2022pit} propose a pyramid-structured Transformer that aggregates information from global to local levels, enabling the extraction of fine-grained features. 
Wu \etal~\cite{wu2024tcvit} introduce a temporal correlation vision Transformer that aligns patch tokens with kernelized correlation filters to enhance the representation of the target person.
While these approaches have achieved remarkable success, they are limited to unimodal frameworks trained solely on image data.
In contrast, we propose a novel vision-language multi-modal learning framework with explicit captions.

\subsection{Vision-language learning}
% The emergence of Vision-Language Models (VLMs), such as CLIP~\cite{radford2021clip}, has significantly advanced visual representation learning by aligning visual features with textual information.
% through the use of massive paired image and text datasets.
% Furthermore, the great development of Large Language Models (LLMs)~\cite{brown2020gpt3, bai2023qwen} has driven the evolution of Multi-modal Large Language Models (MLLMs)~\cite{li2023blip2, xiao2024florence, liu2023llava, abouelenin2025phi4}, which integrate multi-modal information to perform a wide range of tasks.
% In addition, the great development in Large Language Models (LLMs)~\cite{brown2020gpt3, bai2023qwen} has led to the emergence of Multi-modal Large Language Models (MLLMs)~\cite{li2023blip2, xiao2024florence, liu2023llava, abouelenin2025phi4}, which are capable of performing a wide range of tasks.
% For instance, Phi-4-Multimodal (phi-4-mm)~\cite{abouelenin2025phi4} processes three input modalities—text, vision, and speech/audio—using adapters and modality-specific routers, enabling it to handle diverse tasks such as image captioning and speech recognition with high accuracy.

Vision-Language Models (VLMs), such as CLIP~\cite{radford2021clip}, have advanced visual representation learning by aligning visual and textual features.
Recent progress in Large Language Models (LLMs)~\cite{brown2020gpt3, bai2023qwen} has further enabled Multi-modal Large Language Models (MLLMs)~\cite{li2023blip2, xiao2024florence, liu2023llava, abouelenin2025phi4}, which can generate high-quality captions.

Li \etal~\cite{li2023clipreid} apply CLIP to image-based person ReID via a two-stage training scheme, where learnable text tokens are optimized first and then used to fine-tune the image encoder.
Yu \etal~\cite{yu2024tf} build upon this by proposing a one-stage training method for video-based person ReID.
Their approach replaces text features with identity-specific prototypes (CLIP-Memory) extracted from the entire training dataset and employs a Transformer-based module to acquire robust temporal features.
Despite the impressive achievements of these methods, learning with implicit text tokens or averaged image features risks overlooking subtle distinctions between individuals, such as hairstyles or shoe colors.

On the other hand, CLIP-SCGI~\cite{han2024clipscgi} utilizes MLLMs for image-based person ReID, incorporating pseudo-captions into the training pipeline.
%, resulting in remarkable performance improvements over previous methods.
Captions generated for each image are fed into an inversion network to produce text tokens, which are then used for contrastive learning with image features.
However, unlike CLIP-Memory, text tokens are not shared across images of the same identity but are computed per image.
As a result, this approach struggles to capture identity-consistent features.
We propose a novel contrastive learning framework that refines CLIP-Memory using explicit text descriptions for video-based person ReID.

\section{Method}
\label{sec:method}

\begin{figure*}[tbp]
    \centering
    \includegraphics[width=\linewidth]{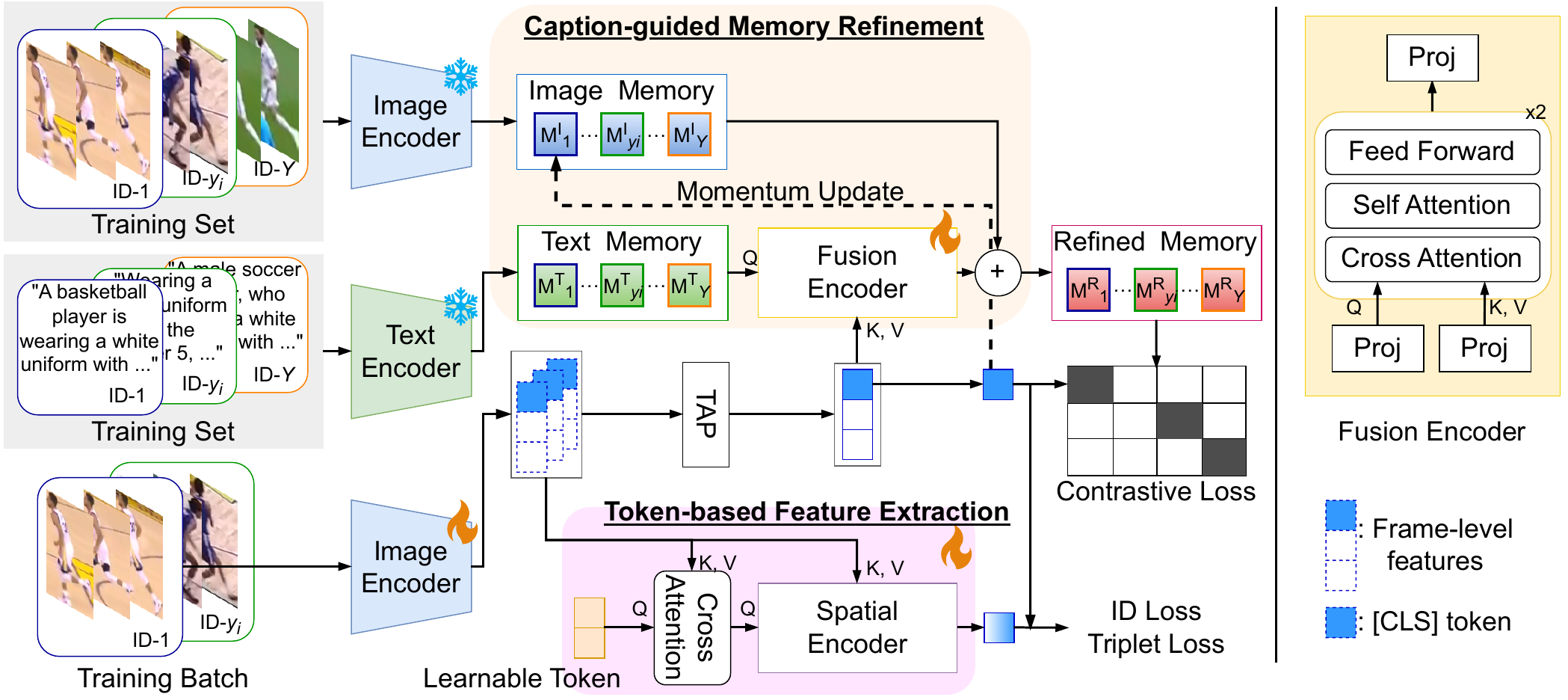}
    \caption{
      Illustration of our proposed CG-CLIP framework.
      Caption-guided Memory Refinement (CMR) refines Image Memory based on synthesized captions to capture fine-grained identity-specific features.
      Token-based Feature Extraction (TFE) aggregates features along both temporal and spatial dimensions using a cross-attention mechanism with learnable tokens.
    }
    \label{fig:method_cmr}
\end{figure*}

\Cref{fig:method_cmr} illustrates the overall architecture of our proposed CG-CLIP framework.
Our method consists of two main modules: Caption-guided Memory Refinement (CMR) and Token-based Feature Extraction (TFE).

\subsection{Preliminaries}
Let a video tracklet containing $L$ images be denoted as $V = \{I_{l}\}^L_{l=1}$.
Here, $I_{l} \in \mathbb{R}^{H \times W \times 3}$ represents the $l$-th image, where $H$ and $W$ indicate the height and width of the image, respectively.
CLIP's image encoder, denoted as $f^I_\theta$, computes deep features from $I_{l}$.
The image is divided into $N^I$ non-overlapping patches and fed into the Transformer blocks following the addition of a [CLS] token.
The output of CLIP's image encoder is given by $F^I_{l} = \{ v^{cls}_{l}; v^{1}_{l}; v^{2}_{l}; \cdots; v^{N^I}_{l}\} \in \mathbb{R}^{(1+N^I)\times D}$, where $v^{cls}_l$ and $v^{n}_l$ represent the $D$-dimensional feature vectors of the class token and patch tokens, respectively.
% Subsequently, the final image feature $F^I_{l} = \{ v^{cls}_{l}; v^{1}_{l}; v^{2}_{l}; \cdots; v^{N^I}_{l}\} \in \mathbb{R}^{(1+N^I)\times D}$ is obtained after passing through the projection layer, where $v^{cls}_l$ and $v^{n}_l$ represent the $D$-dimensional feature vectors of the class token and patch tokens, respectively.
Finally, the class token $v^{cls}_{l}$ is projected to a vision-language unified space via a visual projection layer and Temporal Average Pooling (TAP) is applied to obtain the sequence-level feature $b$.

Text-free methods such as PCL-CLIP~\cite{li2023pcl} and TF-CLIP~\cite{yu2024tf} utilize two image encoders: $f^I_\theta$, whose weights are updated during training, and $f^{I*}_\theta$, whose weights are frozen.
$f^{I*}_\theta$ is used to compute identity-specific image prototypes, referred to as CLIP-Memory.
CLIP-Memory is initialized by inputting the entire training dataset into $f^{I*}_\theta$ and averaging the features for each identity:

\begin{equation}
  M^I_{y_i, init} = \frac{1}{P^I_{y_i}}\sum_{b^* \in y_i} b^*
  \label{eq:Image-Memory}
\end{equation}
Here, $y_i \in \{1, \cdots, Y \}$ represents the label, $P^I_{y_i}$ denotes the total number of sequences belonging to the identity $y_i$, and $b^*$ represents the output from $f^{I*}_\theta$.
Subsequently, CLIP-Memory is updated using the output of $f^I_\theta$ through methods such as momentum update~\cite{li2023pcl, yu2025climb} or Transformer-based update blocks~\cite{yu2024tf}, optimizing it as the centroid for each identity.
The computed memory is then used to calculate the video-to-memory contrastive learning loss:

\begin{equation}
  % L^{(y_i)}_{v2im} = -log \frac{\rm{exp}(s(b_{y_i}, M^I_{y_i}))}{\sum^{B}_{j=1} \rm{exp}(s(b_{y_i}, M^I_j))}
  L^{(y_i)}_{v2im} = \frac{-1}{| D(y_i)|} \sum_{p \in D(y_i)}\log \frac{\text{exp}(s(b_{p}, M^I_{y_i}))}{\sum^{B}_{j=1} \text{exp}(s(b_{j}, M^I_{y_i}))}
  \label{eq:CLIP-Memory_loss}
\end{equation}
Here, $D(y_i)$ denotes the set of positives for $M^I_{y_i}$ in the training batch, and $s()$ represents the cosine similarity function.
$B$ indicates the number of video tracklets in a batch.

In addition to CLIP-Memory $M^I$ (Image Memory), we define an identity-specific Text Memory $M^T$ from captions.
Our goal is to use $M^T$ as a semantic query to select identity-discriminative visual tokens and construct a refined memory target for video-to-memory contrastive learning.

\subsection{Caption-guided Memory Refinement}
In sports and dance scenes, where multiple individuals may be dressed in identical uniforms or costumes, the similarity between persons is extremely high.
Consequently, it is challenging to capture subtle differences using only Image Memory, which is computed based on the average of image features.
To address this, we design a novel module called Caption-guided Memory Refinement (CMR), as illustrated in \cref{fig:method_cmr}, which refines the Memory using MLLM-generated captions to extract fine-grained features.
Captions detailing attributes, such as hairstyles or socks, 
% which is difficult to capture using image-based features alone, 
can effectively emphasize key features essential for person ReID.

First, we input all captions into the pre-trained CLIP text encoder $f^{T*}_\phi$.
This process yields the text features $F^T_j = \{ t^{sos}_j; t^{1}_j; t^{2}_j; \cdots; t^{eos}_j \} \in \mathbb{R}^{N^T\times D}$, where $N^T$ represents the number of text tokens.
Next, the [EOS] token $t^{eos}_j$ is processed through a text projection layer, producing the projected feature $t^{eos'}_j$, which aligns the token to the vision-language space.
Then, we compute the Text Memory $M^T_{y_j}$ by averaging the token $t^{eos'}_j$ for each identity across the entire training dataset, similar to how the Image Memory is initialized:

\begin{equation}
  M^T_{y_j} = \frac{1}{P^T_{y_j}}\sum_{t^{eos'}_j \in y_j} t^{eos'}_{j}
  \label{eq:Text-Memory}
\end{equation}
Here, $P^T_{y_j}$ denotes the total number of text data for label $y_j$.

During training, mini-batches are sampled using the PK sampling strategy~\cite{hermans2017pksample}, with $P$ different identities and $K$ tracklets per identity.
All tokens of the image features $F^I_l$ obtained from the training batch are input into the visual projection layer.
Then, TAP is applied to each token to compute the sequence-level features $\hat{F}^I \in \mathbb{R}^{(1+N^I) \times D}$.
Next, Text Memory $M^T \in \mathbb{R}^{Y \times D}$ and the sequence features $\hat{F}^I$ are input into a fusion encoder.
The transformation is defined as:

\begin{equation}
    \hat{M}^T = FusionEncoder(M^T, \hat{F}^I, \hat{F}^I)
    \label{eq:cmr_transformer_block}
\end{equation}
As illustrated in \cref{fig:method_cmr}, the fusion encoder comprises three projection layers and two Transformer blocks.
Each projection layer is preceded by Layer Normalization (LN).
The Transformer block is composed of a cross-attention layer, a self-attention layer, and a feed-forward network.
Text Memory $M^T$ serves as query, while $\hat{F}^I$ serves as key and value.
Through these computations, an attention map highlights visual tokens that are most relevant to the Text Memory, encouraging the model to focus on identity-discriminative regions.
Finally, we form the Refined Memory $M^R_{y_j}$ by combining the Image Memory $M^I_{y_j}$ with the integrated Text Memory $\hat{M}^T_{y_j}$:

\begin{equation}
    M^R_{y_j} = \hat{M}^T_{y_j} + M^I_{y_j}
\end{equation}
The refined memory $M^R_{y_j}$ incorporates both the representative image features from $M^I_{y_j}$ and the semantic, fine-grained image features guided by captions from $\hat{M}^T_{y_j}$.
Thus, utilizing $M^R_{y_j}$ for contrastive learning can considerably enhance the discriminative performance of the image encoder.

Unlike instance-wise image--text contrastive learning (\eg, CLIP), CMR integrates identity-averaged text features with image features.
This approach stabilizes the Memory and mitigates caption noise/hallucination.
Additionally, the captions and the CMR module are used only during the training phase and do not affect the computational cost or speed during inference.
In our method, the Image Memory is updated using momentum with hard samples~\cite{li2023pcl, yu2025climb}.

\subsection{Token-based Feature Extraction}

\begin{figure}
    \centering
    \includegraphics[width=1.0\linewidth]{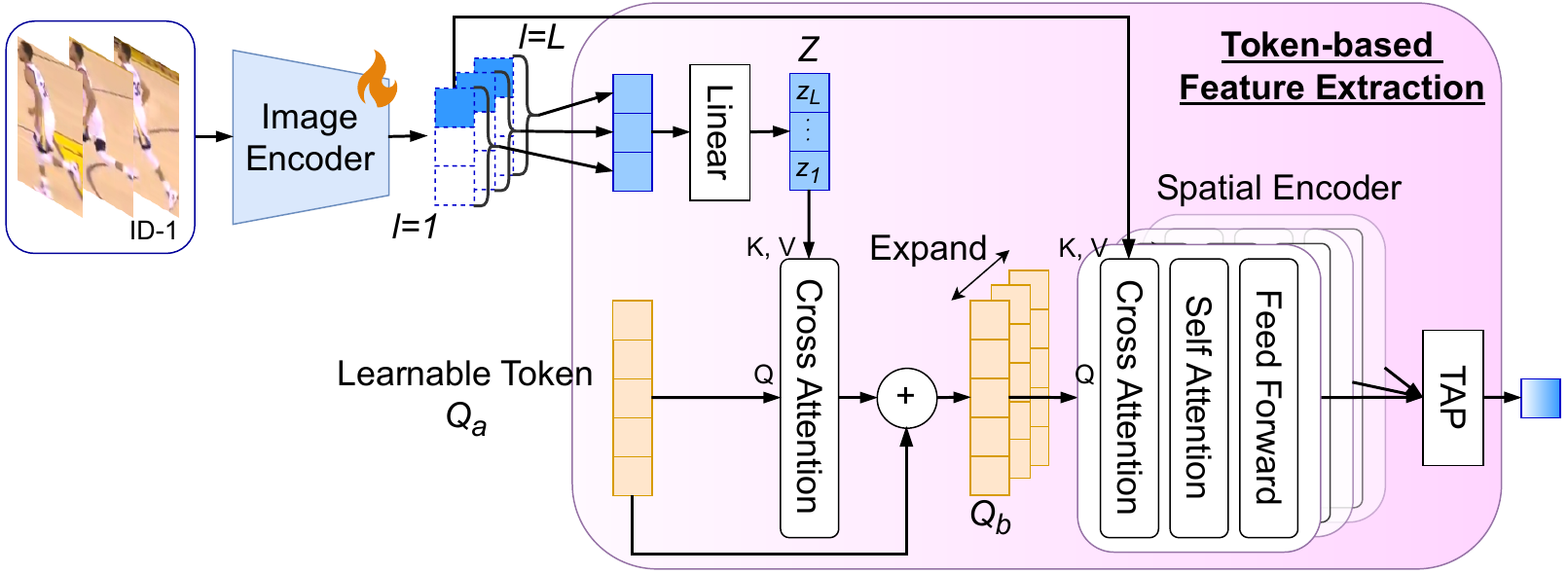}
    \caption{Illustration of the proposed Token-based Feature Extraction (TFE) module.}
    \label{fig:method_tfe}
\end{figure}

Exploiting spatiotemporal features is crucial for video-based person ReID.
Methods such as STFE~\cite{yang2024stfe} and TF-CLIP~\cite{yu2024tf} refine feature maps by applying self-attention mechanisms in both temporal and spatial axes.
However, a significant drawback of self-attention is that its computational cost increases quadratically with the input token length.
In fast-moving sports and dance scenes, it is often necessary to increase the number of frames per tracklet, or process images at higher resolutions, to capture information more densely.
In such scenarios, self-attention-based methods suffer from an increase in computational cost and memory usage.
This limitation can be particularly critical when applying person ReID to real-time tracking scenarios, where high processing speed and efficiency are essential.

To tackle this issue, we propose Token-based Feature Extraction (TFE), as shown in \cref{fig:method_tfe}.
TFE employs fixed-length learnable tokens to linearly constrain the increase in computational cost relative to the input token length.
Consider a tracklet $V = \{I_{l}\}^L_{l=1}$ included in a training batch, where the image features $F^I_{l}$ are obtained through the image encoder.
The TFE module is designed to take in all image tokens $F^I_{l}$ as an input.
We first compute the average of $F^I_{l}$ along the patch dimension and apply a linear projection:

\begin{equation}
    z_l = Linear(\frac{1}{N^I + 1} \sum^{N^I}_{n=0}v^n_l)
\end{equation}
Next, we randomly initialize a set of learnable tokens $Q_a = \{ q_1, q_2, \cdots, q_{N^Q}\} \in \mathbb{R}^{N^{Q} \times D}$ and input them, along with $Z=\{z_1, \cdots, z_L\} \in \mathbb{R}^{L \times D}$, into a single cross-attention layer.
$Q_a$ serves as query, while $Z$ serves as key and value:

\begin{equation}
    Q_b = Q_a + Cross\text{-}Attn(LN(Q_a, Z, Z))
    \label{eq:TFE_temp_attn}
\end{equation}
where $Cross\text{-}Attn$ denotes the cross-attention operation.
Subsequently, $Q_b$ is expanded along the temporal dimension
% to form $Q_c \in \mathbb{R}^{L \times N^{Q} \times D}$
and is then fed into a spatial encoder along with the image features $F^I_l$ of each frame.
The spatial encoder consists of a cross-attention layer, a self-attention layer, and a feed-forward network:

\begin{gather}
    \tilde{Z}_{l} = SpatialEncoder(Q_b, F^I_{l}, F^I_{l})
\end{gather}
where $Q_b$ serves as query, while $F^I_{l}$ serves as key and value.
Finally, we compute the average across all tokens to obtain the frame-level features $\hat{z}_l$ and then apply TAP to derive the sequence-level feature $\hat{b}$:

\begin{gather}
    % \tilde{z_l} = \frac{1}{N^Q}\sum^{N^Q}_{n=1}\hat{z}^n_{l} \\
    \hat{b} = TAP(\{ \hat{z}_1, \cdots, \hat{z}_L\})
\end{gather}
Through the cross-attention mechanism with learnable tokens, the model emphasizes frames and regions crucial for ReID, thereby acquiring robust and distinctive features.

\begin{table}[tbp]
  \caption{
    Overview of the datasets used in our experiments.
    SportsV and DanceV are the newly constructed datasets.
  }
  \centering
  % @{}: delete indent
  \scalebox{0.9}{
  \begin{tabular}{c|cccc}
    \toprule
    Datasets   & SportsV & DanceV &  MARS~\cite{zheng2016mars} & iLIDS~\cite{wang2014ilids} \\
    \midrule
    \# Identities & 513 & 692 & 1,261 & 300 \\
    \# Tracklets & 13,061 & 12,330 & 20,715 & 600 \\
    \# Images    & 599,332 & 574,078 & 1,067,516 & 42,460 \\
    \bottomrule
  \end{tabular}
  }
  \label{tab:dataset}
\end{table}

\subsection{Training and inference}
We employ the video-to-memory contrastive loss with our Refined Memory denoted by $L_{v2rm}$, to train the image encoder and our CMR module.
The loss is defined as follows:
\begin{equation}
  % L^{(y_i)}_{v2rm} = -\log \frac{\text{exp}(s(b_{y_i}, M^R_{y_i}))}{\sum^{B}_{j=1} \text{exp}(s(b_{y_i}, M^R_j))}
  L^{(y_i)}_{v2rm} = \frac{-1}{| D(y_i)|} \sum_{p \in D(y_i)}\log \frac{\text{exp}(s(b_{p}, M^R_{y_i}))}{\sum^{B}_{j=1} \text{exp}(s(b_{j}, M^R_{y_i}))}
  \label{eq:Refined-Memory_loss}
\end{equation}
The encoders for computing Image Memory and Text Memory are frozen during training.
We employ the triplet loss $L_{tri}$~\cite{hermans2017pksample} and the label-smooth cross-entropy loss $L_{ce}$ to update the image encoder and the TFE module. 
Overall, the total loss $L_{total}$ is defined as:
\begin{equation}
    L_{total} = L_{v2rm} + L_{tri} + L_{ce}
\end{equation}

During inference, the output $\hat{b}$ from the TFE module is concatenated with the sequence-level feature $b$ to form the final representation.
% , obtained by applying TAP to the output of the image encoder, to form the final representation.
% Note that textual descriptions and the CMR module are not used in the inference process.

\begin{table*}
  \caption{
    Comparison with state-of-the-art methods across datasets.
    The \textbf{bold} and \underline{underline} denote the best and second results, respectively.
  }
  \centering
  % @{}: delete indent
  \begin{tabular}{ccccccccc}
    \toprule
    \multirow{2}{*}{Methods}   & \multicolumn{2}{c}{MARS} & \multicolumn{2}{c}{iLIDS-VID} & \multicolumn{2}{c}{SportsVReID} & \multicolumn{2}{c}{DanceVReID} \\
      & mAP & Rank-1 & Rank-1 & Rank-5 & mAP & Rank-1 & mAP & Rank-1 \\
    \midrule
    AP3D~\cite{gu2020appearance} & 85.1 & 90.1 & 88.7 & - & 65.4 & 84.2 & 37.9 & 62.0 \\
    TCLNet~\cite{hou2020tclnet} & 85.1 & 89.8 & 86.6 & - & 60.6 & 79.8 & 35.3 & 50.9 \\
    MGH~\cite{yan2020mgh} & 85.8 & 90.0 & 85.6 & 97.1 & 72.6 & \underline{90.1} & 42.9 & 70.1 \\
    GRL~\cite{liu2021grl} & 84.8 & 91.0 & 90.4 & 98.3 & 65.9 & 86.4 & 35.8 & 53.5 \\
    BiCnet-TKS~\cite{hou2021bicnet} & 86.0 & 90.2 & - & - & 62.2 & 80.2 & 34.7 & 47.6 \\
    CTL~\cite{liu2021ctl} & 86.7 & 91.4 & 89.7 & 97.0 & - & - & - & - \\
    STMN~\cite{eom2021stmn} & 84.5 & 90.5 & - & - & - & - & 40.7 & 69.7 \\
    PSTA~\cite{wang2021psta} & 85.8 & 91.5 & 91.5 & 98.1 & 64.7 & 84.2 & 36.8 & 59.4 \\
    % STT~\cite{zhang2021stt} & Arxiv21 & 86.3 & 88.7 & 87.5 & 95.0 & - & - & - & - \\
    PiT~\cite{zang2022pit} & 86.8 & 90.2 & 92.1 & 98.9 & 70.9 & 89.3 & 42.3 & 68.3 \\
    CAVIT~\cite{wu2022cavit} & 87.2 & 90.8 & 93.3 & 98.0 & - & - & - & - \\
    SINet~\cite{bai2022sinet} & 86.2 & 91.0 & 92.5 & - & 73.6 & 88.2 & 42.9 & 70.9 \\
    MFA~\cite{gu2022mfa} & 85.0 & 90.4 & 93.3 & 98.7 & - & - & - & - \\
    DCCT~\cite{liu2023dcct} & 87.5 & 92.3 & 91.7 & 98.6 & 58.9 & 82.4 & 33.8 & 62.4 \\
    LSTRL~\cite{liu2023lstrl} & 86.8 & 91.6 & 92.2 & 98.6 & - & - & - & - \\
    TMT~\cite{liu2024tmt} & 86.5 & 91.8 & 91.3 & 98.6 & - & - & - & - \\
    FT-CLIP~\cite{zhang2024vsla} & 87.8 & 91.5 & 91.3 & 98.7 & 74.4 & 87.5 & 49.7 & 69.7 \\
    VSLA-CLIP~\cite{zhang2024vsla} & 88.2 & 90.9 & \underline{95.3} & - & 74.1 & 89.0 & \underline{52.5} & \underline{73.8} \\
    TCVIT~\cite{wu2024tcvit} & 87.6 & 91.7 & 94.3 & \underline{99.3} & - & - & - & - \\
    % TF-CLIP~\cite{yu2024tf} & AAAI24 & \underline{89.4} & \textbf{93.0} & 94.5 & 99.1 & \underline{76.5} & 89.3 & 49.5 & 66.1 \\
    TF-CLIP~\cite{yu2024tf} & \underline{89.4} & \textbf{93.0} & 94.5 & 99.1 & \underline{77.3} & 89.7 & 51.7 & 70.8 \\
    \textbf{CG-CLIP (Ours)} & \textbf{89.8} & \underline{92.5} & \textbf{96.7} & \textbf{99.9} & \textbf{77.7} & \textbf{90.4} & \textbf{53.8} & \textbf{76.0} \\
    \bottomrule
  \end{tabular}
  \label{tab:SOTA_comparison}
\end{table*}

\subsection{Synthesized caption generation}
As mentioned above, our proposed framework requires both images and text as inputs during training.
However, existing video-based person ReID datasets lack textual data.
% , and manually annotating captions is prohibitively expensive.
To address this limitation, we leverage Multi-modal Large Language Models (MLLMs) to generate captions.
Recently, research on MLLMs has advanced significantly, and many off-the-shelf models have been proposed.
In this work, we utilize Phi-4-Multimodal (phi-4-mm)~\cite{abouelenin2025phi4} to generate effective text for training through multiple steps, such as image captioning, sentence augmentation, and text translation.
Note that for existing benchmark datasets (MARS, iLIDS-VID), captions are generated entirely by phi-4-mm, whereas for our proposed datasets (SportsVReID, DanceVReID), captions are generated by phi-4-mm based on manually annotated attributes obtained during dataset creation.
% For further details, refer to the supplementary material Sec.~\ref{sec:suppl_caption_generation}.

\section{Experiments}
\label{sec:experiments}

\subsection{Datasets and evaluation protocols}
We evaluate our method on two existing video-based person ReID datasets, MARS~\cite{zheng2016mars} and iLIDS-VID~\cite{wang2014ilids}, as well as two newly constructed high-difficulty datasets, SportsVReID and DanceVReID, derived from Multi-Object Tracking (MOT) benchmarks (SportsMOT~\cite{cui2023sportsmot} and DanceTrack~\cite{sun2022dancetrack}).
% To transform these MOT datasets into ReID datasets, we first utilize the person labels and bounding box information provided in each MOT dataset to crop person regions from each frame.
For SportsVReID/DanceVReID, we first crop person images using MOT labels and bounding boxes.
Subsequently, we employ annotators to describe the characteristics of each person (\eg, gender, hairstyle, clothing, socks, shoes, and jersey number).
Based on these annotations, we link the same individuals appearing in different videos to a single label and then divide the videos into tracklets (short video clips).
We follow the original train/val splits; for evaluation, the first tracklet per identity is used as query and the rest as gallery.
\Cref{tab:dataset} provides a brief summary of each dataset (SportsV/DanceV/iLIDS denote SportsVReID/DanceVReID/iLIDS-VID).
% See the supplementary material Sec.~\ref{sec:suppl_dataset_creation} and Sec.~\ref{sec:suppl_dataset_visualization} for details.
We adopt Cumulative Matching Characteristic (CMC) at Rank-k and mean Average Precision (mAP) as evaluation metrics. 

\subsection{Experiment settings}
We use the CLIP image encoder (ViT-B/16) and text encoder.
During training, we sample 8 images from each video tracklet and resize them to $256\times128$.
For data augmentation, we adopt random flipping and random erasing~\cite{zhong2020randomerase}.
The maximum length of the input captions is set to 77 tokens.
The batch size is 32, containing 8 identities with 4 tracklets per identity.
We train our framework for 80 epochs on MARS and 60 epochs on the other datasets.
We use the Adam optimizer with a learning rate of $5 \times 10^{-6}$, linearly warming up the model from $5 \times 10^{-7}$ over the first 10 epochs.
Subsequently, we reduce the learning rate by a factor of 10 at the 30th, 50th, and 70th epochs for MARS, and at the 30th and 50th epochs for the other datasets.
The number of learnable tokens ($N^Q$) in TFE is set to 50 for MARS and 15 for the other datasets.
We use cosine distance for triplet loss and similarity calculation during inference.
% Our model is implemented with PyTorch.

\begin{table} % [htbp]
  % \begin{minipage}[c]{0.65\hsize}
  \caption{Comparison of different components.}
  \centering
  % @{}: delete indent
  \scalebox{0.92}{
  \begin{tabular}{ccccccc}
    \toprule
    \multirow{2}{*}{Model} & \multicolumn{2}{c}{Components} & \multicolumn{2}{c}{MARS} & \multicolumn{2}{c}{SportsVReID} \\
      & CMR & TFE & mAP & Rank-1 & mAP & Rank-1 \\
    \midrule
    1 & $\times$ & $\times$ & 88.5 & 91.2 & 75.5 & 88.6 \\
    2 & \checkmark &$\times$ & 89.7 & 92.2 & 77.3 & 87.9 \\
    3 & $\times$ & \checkmark & 88.5 & 91.5 & 76.6 & 89.0 \\
    4 & \checkmark & \checkmark & 89.8 & 92.5 & 77.7 & 90.4 \\
    \bottomrule
  \end{tabular}
  }
  \label{tab:ablation_components}
  % \end{minipage}
\end{table}

\begin{table}[tbp]
  % \begin{minipage}[c]{0.35\hsize}
  \caption{Comparison of different types of Memory.}
  \centering
  % @{}: delete indent
  \small
  \setlength{\tabcolsep}{4pt}
  \scalebox{0.96}{
  \begin{tabular}{ccccc}
    \toprule
    \multirow{2}{*}{Memory} & \multicolumn{2}{c}{MARS} & \multicolumn{2}{c}{SportsVReID} \\
      & mAP & Rank-1 & mAP & Rank-1 \\
    \midrule
    Image Mem. only & 88.5 & 91.5 & 76.6 & 89.0 \\
    Text Mem. only & 86.8 & 90.6 & 73.1 & 84.6 \\
    Image+Text Mem. (naive sum) & 89.3 & 92.2 & 77.2 & 87.9 \\
    Refined Mem. (ours, w/ CMR) & 89.8 & 92.5 & 77.7 & 90.4 \\
    \bottomrule
  \end{tabular}
  }
  \label{tab:ablation_memory}
  % \end{minipage}
\end{table}

\subsection{Comparison with state-of-the-art methods}
Our results are presented in \cref{tab:SOTA_comparison}, demonstrating that our method outperforms previous methods on both existing datasets and our new challenging datasets.

\thirdheading{Results on existing datasets}
Our method achieves the best mAP of 89.8\% and a Rank-1 92.5\% on MARS, outperforming many other approaches.
Specifically, our method surpasses FT-CLIP~\cite{zhang2024vsla}, which extends CLIP-ReID~\cite{li2023clipreid} by using TAP to tackle the video domain, by 2.0\% in mAP and 1.0\% in Rank-1.
For iLIDS-VID, we reach 96.7\% Rank-1 accuracy, which is 5.4\% higher than TMT~\cite{liu2024tmt}, which aggregates multi-view features.

\thirdheading{Results on new datasets}
On SportsVReID, we achieve the best mAP of 77.7\% and Rank-1 of 90.4\%, surpassing TF-CLIP~\cite{yu2024tf} by 0.4\% in mAP and 0.7\% in Rank-1.
The difference is greater on DanceVReID, where our method outperforms TF-CLIP by 5.2\% in Rank-1.
While TF-CLIP fine-tunes the CLIP image encoder with Image Memory, our approach further optimizes the model using text captions to emphasize differences between individuals.
This demonstrates the superiority of our approach on high-difficulty scenarios where individuals wear nearly identical uniforms.

\subsection{Ablation study}
We conduct additional experiments to further investigate the impact and effectiveness of each component.
The results are presented in \cref{tab:ablation_components}.
\textit{Model 1} refers to the baseline model that performs contrastive learning using only Image Memory without captions and aggregates features using TAP.

\thirdheading{Effectiveness of CMR}
As shown in the first two rows of \cref{tab:ablation_components}, \textit{Model 2}, which introduces CMR, outperforms \textit{Model 1} on MARS by 1.2\% in mAP and 1.0\% in Rank-1.
On SportsVReID, while the Rank-1 accuracy is almost identical, a 1.8\% improvement in mAP is observed.
These results indicate that incorporating captions refines the memory, enhancing the effectiveness of contrastive learning.

\thirdheading{Effectiveness of TFE}
We further verify the impact of TFE, as shown in \cref{tab:ablation_components}.
Compared with \textit{Model 1}, \textit{Model 3}, which incorporates the TFE module, achieves a 0.3\% improvement in Rank-1 on MARS and a 1.1\% gain in mAP on SportsVReID. 
Furthermore, when comparing \textit{Model 2} and \textit{Model 4}, we observe a 0.3\% increase in Rank-1 on MARS and a 2.5\% increase in Rank-1 on SportsVReID.
These results indicate that the improvement is particularly significant on SportsVReID, suggesting that the extraction of temporal and spatial information is especially beneficial in dynamic sports scenes compared to pedestrian scenarios.
% , where movements are more intense compared to pedestrian scenarios.

\thirdheading{Comparison of different types of Memory}
% To justify the design of CMR, we compare different ways of constructing identity prototypes in \cref{tab:ablation_memory}.
\Cref{tab:ablation_memory} compares different ways of constructing identity prototypes.
A naive fusion (Image+Text Memory, simple summation) improves mAP over Image Memory only but degrades Rank-1 on SportsVReID, suggesting that directly mixing image and text prototypes may introduce noisy cues in high-similarity scenarios.
In contrast, our Refined Memory achieves the best overall performance, demonstrating that CMR is not a trivial combination of image and text prototypes but a caption-guided refinement that selectively injects identity-discriminative cues into the memory target.

\begin{table}[tbp]
  % \begin{minipage}[c]{0.35\hsize}
  \caption{Effect of the number of learnable tokens in TFE.}
  \centering
  % @{}: delete indent
  \scalebox{0.92}{
  \begin{tabular}{ccccc}
    \toprule
    \# Token & \multicolumn{2}{c}{MARS} & \multicolumn{2}{c}{SportsVReID} \\
    ($N^Q$)  & mAP & Rank-1 & mAP & Rank-1 \\
    \midrule
    5 & 89.5 & 92.1 & 77.1 & 89.3 \\
    % 10 & 89.8 & 92.2 & 77.8 & 89.7 \\
    15 & 89.6 & 92.2 & 77.7 & 90.4 \\
    % 30 & 89.5 & 92.0 & 77.3 & 89.0 \\
    50 & 89.8 & 92.5 & 77.8 & 89.3 \\
    % 70 & 89.7 & 92.4 & 77.4 & 89.7 \\
    100 & 89.6 & 92.4 & 77.0 & 88.2 \\
    % 150 & 89.4 & 91.9 & 77.3 & 90.1 \\
    200 & 89.4 & 92.6 & 77.2 & 88.2 \\
    \bottomrule
  \end{tabular}
  }
  \label{tab:ablation_token}
  % \end{minipage}
\end{table}

\begin{table*}[tbp]
  \caption{
    Comparison under varying training data sizes on MARS and SportsVReID (mAP).
  }
  \centering
  % @{}: delete indent
  \scalebox{0.95}{
  \begin{tabular}{c|ccccc|ccccc}
    \toprule
    \multirow{2}{*}{Methods} & \multicolumn{5}{c|}{MARS} & \multicolumn{5}{c}{SportsVReID} \\
      & 20\% & 40\% & 60\% & 80\% & 100\% & 20\% & 40\% & 60\% & 80\% & 100\% \\
    \midrule
    FT-CLIP~\cite{zhang2024vsla} & 78.2 & 82.2 & 84.6 & 87.0 & 87.8 & 63.9 & 68.6 & 70.8 & 74.7 & 74.4 \\
    VSLA-CLIP~\cite{zhang2024vsla} & 78.1 & 83.2 & 86.1 & 87.7 & 88.2 & 57.8 & 68.6 & 70.2 & 73.7 & 74.1 \\
    TF-CLIP~\cite{yu2024tf} & 79.5 & 84.1 & 87.0 & 88.1 & 89.4 & 65.4 & 70.5 & 73.3 & \textbf{76.3} & 77.3 \\
    \textbf{CG-CLIP (Ours)} & \textbf{81.1} & \textbf{85.6} & \textbf{87.7} & \textbf{88.7} & \textbf{89.8} & \textbf{66.6} & \textbf{72.8} & \textbf{74.2} & \textbf{76.3} & \textbf{77.7} \\
    \bottomrule
  \end{tabular}
  }
  \label{tab:data_efficiency}
\end{table*}

\thirdheading{Effectiveness of the number of learnable tokens}
We conduct experiments to evaluate the effect of changing the number of learnable tokens in TFE.
As shown in \cref{tab:ablation_token}, it is evident that our method is not highly sensitive to this hyper-parameter.
To balance computation and performance, we set $N^{Q} = 50$ for MARS and $N^{Q} = 15$ for the other datasets in our default setting.

\begin{figure}[tbp]
    \centering
    \begin{subfigure}{0.235\textwidth}
        \centering
        \includegraphics[width=\textwidth]{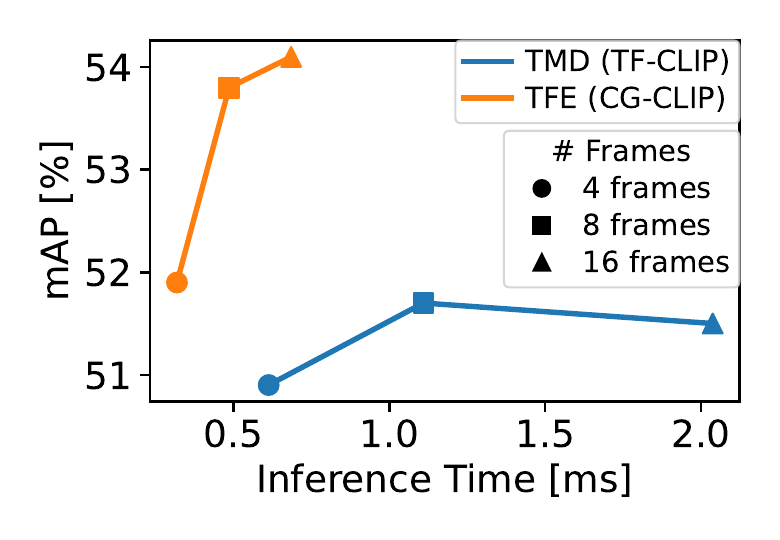}
        \caption{Frames.}
        \label{fig:speed_frame}
    \end{subfigure}
    \hfill
    \begin{subfigure}{0.235\textwidth}
        \centering
        \includegraphics[width=\textwidth]{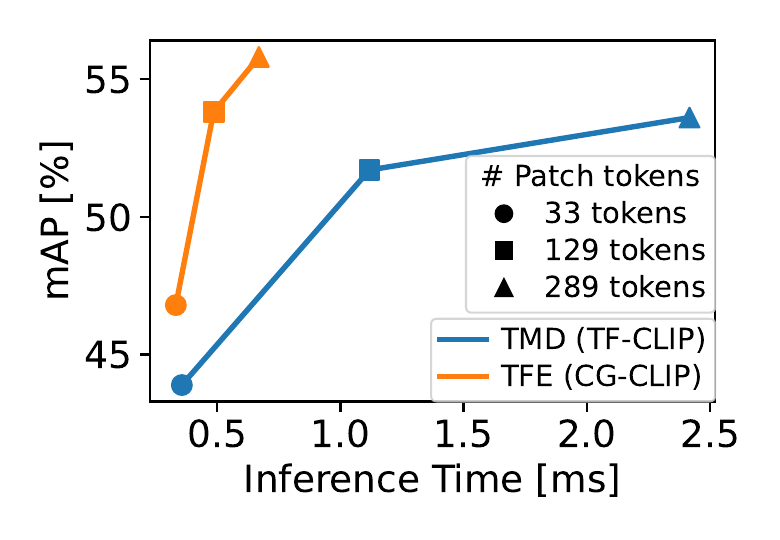}
        \caption{Patch tokens.}
        \label{fig:speed_token}
    \end{subfigure}
    \caption{Accuracy--latency trade-off on DanceVReID}
    \label{fig:tfe_scalability}
\end{figure}

\thirdheading{Comparison of data efficiency}
We evaluate data efficiency on MARS and SportsVReID by training with 20--100\% of the training set.
We subsample identities (not tracklets) to keep all tracklets of each selected identity.
As shown in \cref{tab:data_efficiency}, our method consistently outperforms all baseline methods across all data regimes on both datasets.
Notably, the performance gap becomes more pronounced when training data is limited.
For instance, on SportsVReID with 20\% training data, our method outperforms TF-CLIP by 1.2\% in mAP.
These results demonstrate that explicit caption guidance enables more efficient learning with limited data, as captions provide rich semantic information that helps the model capture discriminative features.
% even when visual training samples are scarce.

\thirdheading{Comparison of inference speed}
To verify the speed advantage of our proposed method, we conduct experiments on the temporal aggregation module.
We compare our TFE (with $N^{Q}=15$) against Temporal Memory Diffusion (TMD)~\cite{yu2024tf}, which integrates temporal and spatial information via self-attention.
Both methods are converted from PyTorch models to TensorRT and deployed on an NVIDIA RTX A4000 GPU.
We measure the inference speed of each module while varying the number of input frames and the input resolution (\ie, the number of patch tokens).
For accuracy, we train a separate model on DanceVReID for each input setting.
\Cref{fig:tfe_scalability} shows a better accuracy--latency trade-off than TMD when scaling frames/patch tokens.
Compared to TMD, at $L$=8 and 256$\times$128, TFE reduces temporal-module compute from 7.6 to 2.3 GMACs ($\sim$70\%) with a small parameter increase (10.0M$\rightarrow$12.4M).

\subsection{Visualization}
\begin{figure}
    \centering
    \includegraphics[width=1.0\linewidth]{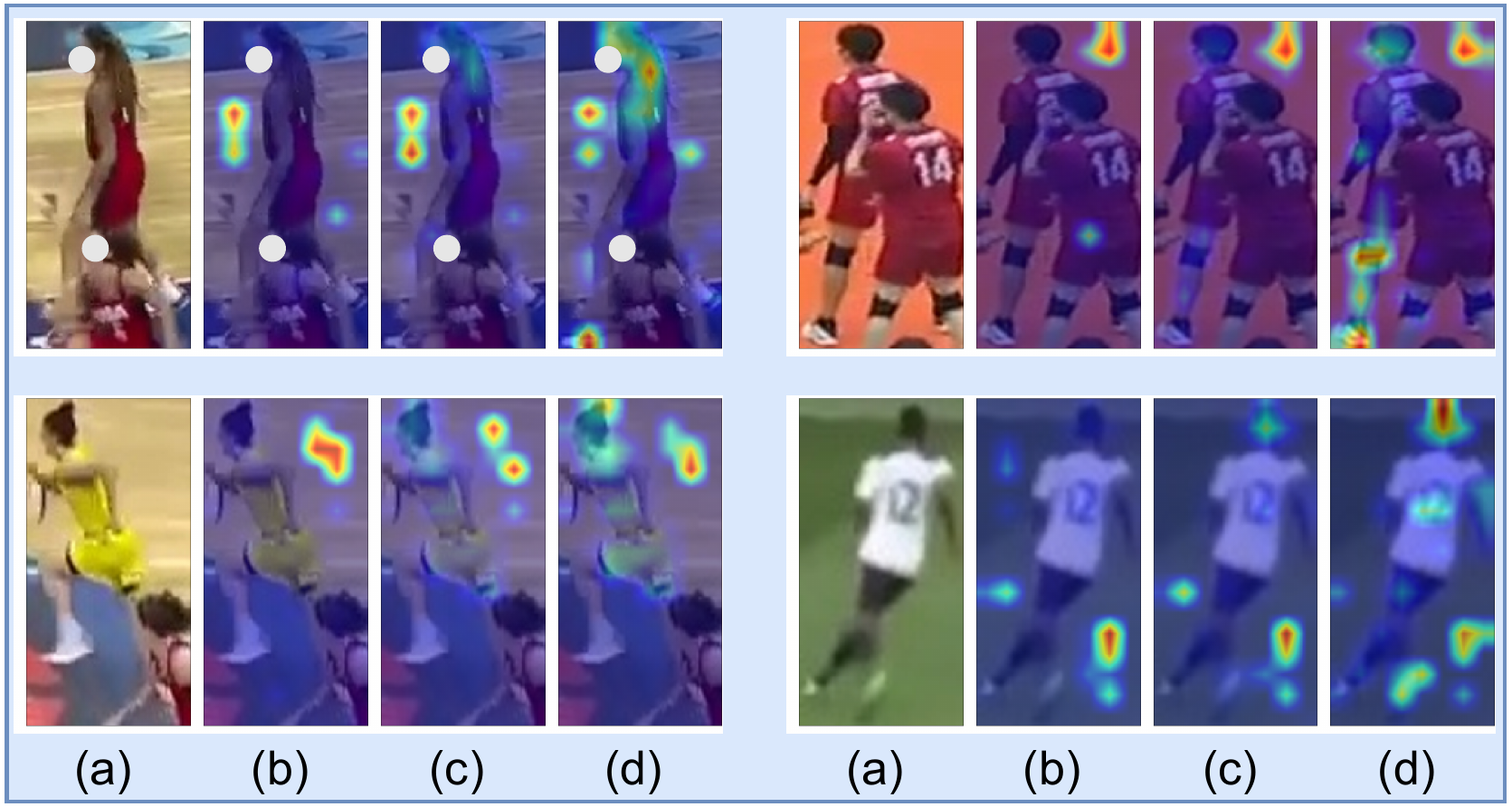}
    \caption{
          Visualization of attention maps on SportsVReID.
          (a) Original images, (b) Vanilla CLIP, (c) TF-CLIP, (d) CG-CLIP.
          Our method effectively focuses on fine-grained, discriminative regions.
    }
    \label{fig:vit_attention}
\end{figure}
To investigate which regions our model focuses on, we visualize the attention maps between the [CLS] token and patch tokens in the final layer of the image encoder on SportsVReID, compared with Vanilla CLIP and TF-CLIP.
As shown in \cref{fig:vit_attention}, our method exhibits stronger attention on discriminative human body regions that are crucial for person ReID, such as hairstyles, shoes, and jersey numbers, compared to the other methods.
This demonstrates that our caption-guided mechanism enables the model to focus more effectively on fine-grained features, which are often overlooked by methods relying solely on image-based learning.

\section{Conclusion}
\label{sec:conclusion}
In this paper, we address a critical challenge in video-based person ReID: identifying individuals in high-difficulty scenarios where multiple people wear nearly identical clothing. 
To tackle this problem, we propose CG-CLIP, a novel caption-guided CLIP framework with two key technical contributions.
First, our Caption-guided Memory Refinement (CMR) module leverages MLLM-generated captions to enhance discriminative performance by integrating textual descriptions with image features.
Second, our Token-based Feature Extraction (TFE) module efficiently aggregates spatiotemporal information through a cross-attention mechanism with learnable tokens, achieving robust representations while maintaining computational efficiency.
Additionally, we construct new benchmark datasets (SportsVReID and DanceVReID) featuring sports and dance performance scenarios that reflect real-world high-similarity challenges.
Extensive experiments show consistent improvements over state-of-the-art methods on both existing and newly proposed datasets.

\thirdheading{Acknowledgments} We are grateful to Tokuhiro Nishikawa and Helen Suzuki for reviewing this manuscript.

{
    \small
    \bibliographystyle{ieeenat_fullname}
    \bibliography{main}
}

\clearpage
\setcounter{page}{1}
\maketitlesupplementary

% ===== Supplementary numbering =====
\appendix

% Sections: A, B, C, ...
\renewcommand{\thesection}{\Alph{section}}
\renewcommand{\thesubsection}{\thesection.\arabic{subsection}}

% Figures/Tables: S.1, S.2, ...
\renewcommand{\thefigure}{S.\arabic{figure}}
\renewcommand{\thetable}{S.\arabic{table}}
\setcounter{figure}{0}
\setcounter{table}{0}

% (optional) Equations: S.1, S.2, ...
\renewcommand{\theequation}{S.\arabic{equation}}
\setcounter{equation}{0}
% ===================================

% \stepcounter{section}
% \stepcounter{section}
% \stepcounter{section}
% \stepcounter{section}
% \stepcounter{section}
\section{Dataset}
\label{sec:suppl_dataset}

\subsection{Source MOT dataset}
To evaluate video-based person ReID methods in challenging scenarios with high visual similarity between individuals, we construct two new datasets.
We focus on Multi-Object Tracking (MOT) datasets, as they provide person ID labels for each individual along with their bounding boxes in each frame.
We adopt the SportsMOT dataset~\cite{cui2023sportsmot} and the DanceTrack dataset~\cite{sun2022dancetrack}.
The SportsMOT dataset contains 240 video clips collected from three sports: basketball, soccer, and volleyball.
The DanceTrack dataset comprises 100 video clips capturing group dance scenes.  
Both datasets feature multiple individuals wearing similar clothing, making it extremely challenging to distinguish each person.

\subsection{Dataset creation process}
\label{sec:suppl_dataset_creation}
MOT datasets typically include videos, per-frame person ID labels, and bounding boxes.
First, using the person ID labels and bounding boxes, we crop the regions corresponding to each individual from every frame of each video.
Since the labels in MOT datasets are assigned independently for each video and do not correspond across videos, it is not possible to automatically detect and match the same individual appearing in different videos.
Therefore, we employ two annotators to describe the characteristics of each individual (\eg, gender, hairstyle, clothing, socks, shoes, and jersey number) for every video.
\Cref{tab:suppl_anno_example} shows examples of the annotations provided in our SportsVReID dataset.
Note that the actual annotations are originally created in Japanese and translated into English for this paper.
Based on these descriptions, we manually perform person matching across videos and reassign new labels to the entire dataset.

\begin{table}[htbp]
  % \begin{minipage}[c]{0.35\hsize}
  \caption{Examples of manual annotations for SportsVReID.}
  \centering
  % @{}: delete indent
  \scalebox{0.85}{
  \begin{tabular}{cccccc}
    \toprule
    ID & Sports & Gender & Uniform & Number & Others \\
    \midrule
    1 & basketball & woman & yellow & 5 & bun hair \\
    2 & basketball & woman & blue & 10 & red shoes \\
    91 & soccer & man & gray, blue & 32 & black socks \\
    92 & soccer & man & white & 1 & red shoes \\
    \bottomrule
  \end{tabular}
  }
  \label{tab:suppl_anno_example}
  % \end{minipage}
\end{table}

Subsequently, we filter out images where the target individual is barely visible or the image size is too small.
We then divide the videos into tracklets (short video clips) for each person, ensuring that the maximum frame length is 50.
Through this process, we create the SportsVReID and DanceVReID datasets from the SportsMOT and DanceTrack datasets, respectively.
It is worth noting that each MOT dataset is divided into three subsets: train, val, and test.
However, since the test set does not include ground truth (\ie, no labels or bounding box information), we use only the train and val sets, with the train set for training and the val set for evaluation.

\subsection{Caption generation}
\label{sec:suppl_caption_generation}
In this paper, we utilize Multi-modal Large Language Models (MLLMs), such as Phi-4-Multimodal (phi-4-mm)~\cite{abouelenin2025phi4}, to perform image captioning, caption augmentation, and translation for generating captions.
For each task, we first input a few examples into GPT-4o~\cite{hurst2024gpt4o} to generate response examples, which are then included in the prompt to conduct caption generation in a few-shot manner.

\thirdheading{Captions for existing datasets}
For existing benchmark datasets (MARS~\cite{zheng2016mars}, iLIDS-VID~\cite{wang2014ilids}), we generate one caption per image using phi-4-mm.
Below, we present the prompt sample used for the image captioning task:
\begin{quote}
\textit{Write a description about the overall appearance of the person in the image, including the attributes: clothes, shoes, hairstyle, gender, belongings.}

\textit{(Output Examples)}
\begin{itemize}
    \item \textit{A man is wearing a white short-sleeved T-shirt and black long pants. His shoes are gray and he has short hair.}
    \item \textit{A woman with long black hair is dressed in a pink short-sleeved shirt and short navy blue pants. She is wearing pink sandals.}
\end{itemize}
\end{quote}
% Specifically for MARS, which is relatively larger in scale than other datasets, we perform caption-to-caption generation to increase the variety of captions.
Additionally, for the MARS dataset, which is relatively larger in scale compared to other datasets, we perform caption-to-caption generation to increase caption diversity.
This process creates multiple diverse captions with semantically equivalent content from a single source caption.

\begin{table*}[htbp]
  \caption{
    Examples of images and captions from the SportsVReID and DanceVReID datasets.
    Each row corresponds to a different person.
  }
  \centering
  \begin{tabular}{|c|c|p{9cm}|}
    \hline
    Dataset & Images & \multicolumn{1}{c|}{Caption} \\
    \hline
     & 
    \begin{minipage}[c][2.1cm][c]{4truecm}
      \centering
      \includegraphics[height=2cm]{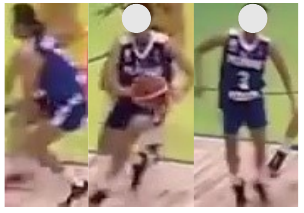}
    \end{minipage} & 
    \begin{minipage}[c][2.1cm][c]{9truecm}
      A female basketball player is wearing a blue uniform with the number 3. She has a ponytail.
    \end{minipage} \\
    \cline{2-3}
    SportsVReID & 
    \begin{minipage}[c][2.1cm][c]{4truecm}
      \centering
      \includegraphics[height=2cm, keepaspectratio]{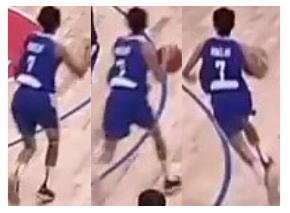}
    \end{minipage} & 
    \begin{minipage}[c][2.1cm][c]{9truecm}
      A woman basketball player is seen in a blue uniform with the number 7, and she also wears black shoes.
    \end{minipage} \\
    \cline{2-3}
    & 
    \begin{minipage}[c][2.1cm][c]{4truecm}
      \centering
      \includegraphics[height=2cm, keepaspectratio]{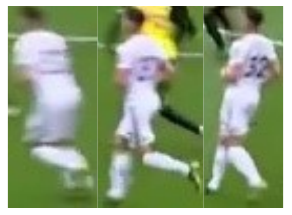}
    \end{minipage} & 
    \begin{minipage}[c][2.1cm][c]{9truecm}
      A man soccer player, in a white uniform with the number 32, also has black shoes.
    \end{minipage} \\
    \hline
     & 
    \begin{minipage}[c][2.1cm][c]{4truecm}
      \centering
      \includegraphics[height=2cm, keepaspectratio]{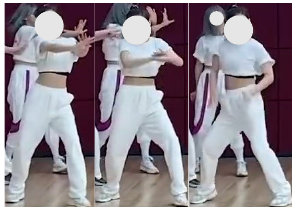}
    \end{minipage} & 
    \begin{minipage}[c][2.1cm][c]{9truecm}
      A female is dressed in a white cropped T-shirt with a black inner layer and white track pants. She wears white sneakers and her hair is styled in a half-updo.
    \end{minipage} \\
    \cline{2-3}
    DanceVReID & 
    \begin{minipage}[c][2.1cm][c]{4truecm}
      \centering
      \includegraphics[height=2cm, keepaspectratio]{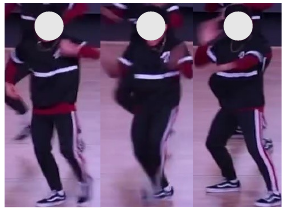}
    \end{minipage} & 
    \begin{minipage}[c][2.1cm][c]{9truecm}
      A male is wearing a shirt with red, white, and black stripes on the upper body and red, white, and black striped pants on the lower body. He also has on white socks and white sneakers with black accents, and his hair is short.
    \end{minipage} \\
    \cline{2-3}
    & 
    \begin{minipage}{4truecm}
      \centering
      \includegraphics[height=2cm, keepaspectratio]{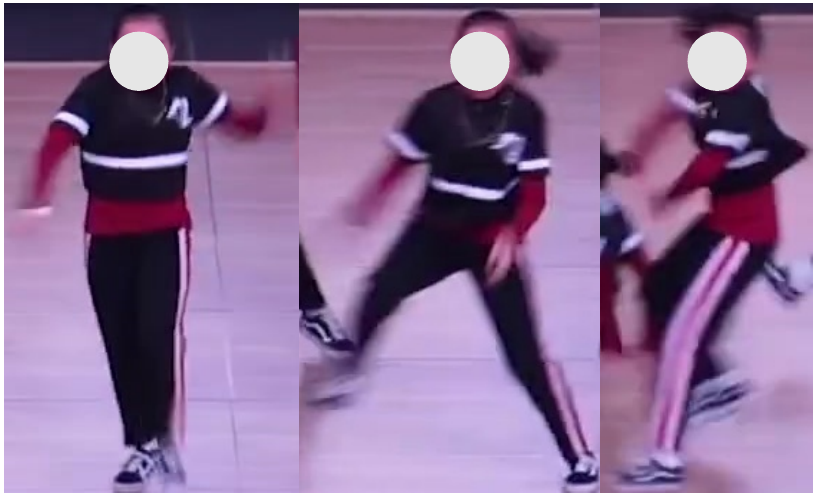} 
    \end{minipage} & 
    \begin{minipage}[c][2.1cm][c]{9truecm}
      She is a woman dressed in a red, white, and black top and matching red, white, and black pants. She also wears white socks and white sneakers with black accents, and her hair is in a ponytail. She is wearing a short jacket.
    \end{minipage} \\
    \hline
  \end{tabular}
  \label{tab:dataset_example}
\end{table*}

\thirdheading{Captions for new datasets}
For our SportsVReID and DanceVReID datasets, we synthesize captions based on manually assigned annotations created during the dataset creation process.
First, the annotation data written in Japanese is translated into English using phi-4-mm.
Since the annotation data contain only one description per identity, we perform paraphrasing with phi-4-mm to create variations of the translated sentences.
This process expands the data to 10 captions per identity.

\subsection{Dataset visualization}
\label{sec:suppl_dataset_visualization}
\Cref{tab:dataset_example} shows examples from our SportsVReID and DanceVReID datasets.
Each row displays a video sequence of a different person along with the corresponding caption. 
Both SportsVReID and DanceVReID include individuals wearing nearly identical uniforms or costumes, making them significantly more challenging person ReID datasets than previously available benchmarks (\eg, MARS).

\section{Ablation study}
\label{sec:suppl_ablation}

\subsection{Analysis of the fusion encoder in CMR}
\begin{figure}[htbp]
    \centering
    \includegraphics[width=1.0\linewidth]{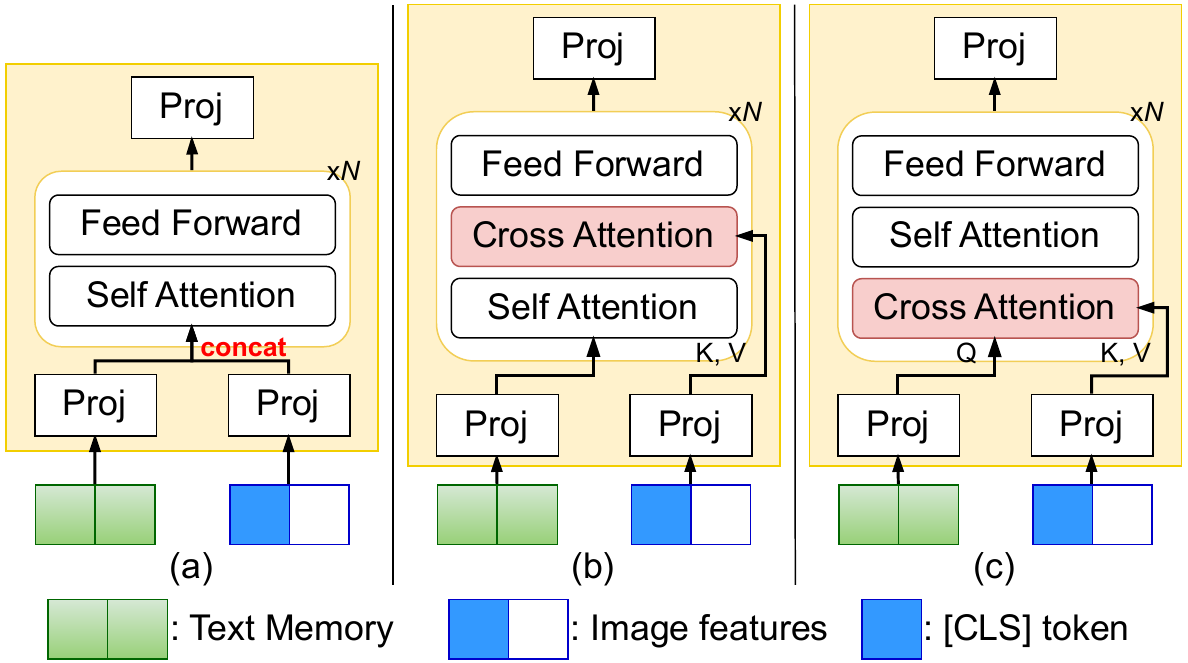}
    \caption{Illustration of three types of fusion encoders in CMR.}
    \label{fig:cmr_fusion}
\end{figure}

We investigate the effectiveness of different approaches for processing the Text Memory and image features in the Caption-guided Memory Refinement (CMR) module.
As illustrated in \cref{fig:cmr_fusion}, we evaluate three configurations:
(a) concatenating Text Memory and image features before feeding them into a self-attention layer,
(b) applying a self-attention layer to the Text Memory before feeding it into the cross-attention layer with image features, where image features serve as keys and values,
and (c) our proposed method, which first applies a cross-attention layer with image features as keys and values, followed by a self-attention layer.
All configurations use 2 Transformer blocks.
As observed in \cref{tab:ablation_cgm_fusion}, our proposed method (c) outperforms the other two configurations on both the MARS and SportsVReID datasets.

\begin{table}[tbp]
  % \begin{minipage}[c]{0.35\hsize}
  \caption{Comparison of different types of fusion encoders in CMR.}
  \centering
  \scalebox{0.88}{
  % @{}: delete indent
  \begin{tabular}{c|cccc}
    \toprule
    \multirow{2}{*}{Method} & \multicolumn{2}{c}{MARS} & \multicolumn{2}{c}{SportsVReID} \\
      & mAP & Rank-1 & mAP & Rank-1 \\
    \midrule
    (a) Self-attn (concat) & 89.6 & 92.0 & 76.9 & 87.1 \\
    (b) Self-attn $\rightarrow$ Cross-attn & 89.7 & 92.4 & 77.2 & 88.2 \\
    (c) Cross-attn $\rightarrow$ Self-attn & 89.8 & 92.5 & 77.7 & 90.4 \\
    \bottomrule
  \end{tabular}
  }
  \label{tab:ablation_cgm_fusion}
  % \end{minipage}
\end{table}

We further analyze the impact of varying the number of Transformer blocks in the CMR module.
As shown in \cref{tab:ablation_cgm_layer}, we evaluate configurations with 1, 2, 3, and 4 blocks on MARS and SportsVReID.
The results indicate that using 2 blocks achieves the best performance across both datasets, attaining the highest Rank-1 scores.
Increasing the number of blocks beyond 2 does not consistently improve performance; an excessive number of blocks (\eg, 3 or 4) may lead to slight performance degradation, likely due to overfitting or increased model complexity.

\begin{table}[tbp]
  % \begin{minipage}[c]{0.35\hsize}
  \caption{Effect of varying the number of Transformer blocks in CMR.}
  \centering
  % @{}: delete indent
  \begin{tabular}{ccccc}
    \toprule
    \multirow{2}{*}{\# Blocks} & \multicolumn{2}{c}{MARS} & \multicolumn{2}{c}{SportsVReID} \\
      & mAP & Rank-1 & mAP & Rank-1 \\
    \midrule
    1 & 89.5 & 91.9 & 77.5 & 90.1 \\
    2 & 89.8 & 92.5 & 77.7 & 90.4 \\
    3 & 89.6 & 91.9 & 76.8 & 87.1 \\
    4 & 89.6 & 92.4 & 77.8 & 88.2 \\
    \bottomrule
  \end{tabular}
  \label{tab:ablation_cgm_layer}
  % \end{minipage}
\end{table}

\subsection{Identity-aware text strategies}
\begin{figure}[tbp]
    \centering
    \includegraphics[width=1.0\linewidth]{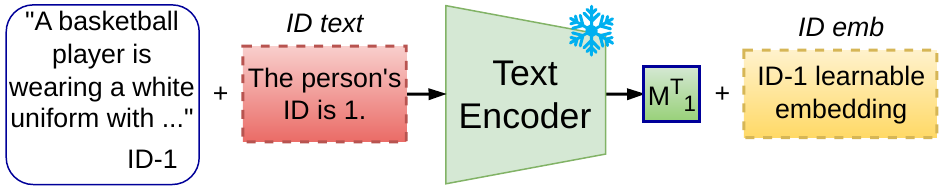}
    \caption{Illustration of text variants used to construct the Text Memory on existing benchmarks (MARS and iLIDS-VID): caption, \textit{ID text}, and \textit{ID emb}.}
    \label{fig:id_text_types}
\end{figure}

For existing benchmark datasets, automatically generated captions may suffer from hallucination effects, where MLLMs assign the same or highly similar captions to different individuals.
This poses a critical issue for our method, as we use features derived from captions as targets for contrastive learning, which requires these features to be unique to each identity.
To encourage identity-unique text features, we investigate two simple strategies summarized in \cref{fig:id_text_types} on MARS and iLIDS-VID, where captions are fully generated by an MLLM.
\textbf{(1) ID text.} We append a short identity string to each caption: \textit{``The person's ID is [ID LABEL].''}
\textbf{(2) ID emb.} We add a learnable identity embedding to the caption feature, analogous to the positional embeddings in Transformers.
\Cref{tab:ablation_text} shows that combining captions with \textit{ID text} yields the best performance on both datasets.
This suggests that explicitly injecting identity information is an effective and lightweight way to mitigate caption ambiguity on existing benchmarks.

\begin{table}[tbp]
  % \begin{minipage}[c]{0.65\hsize}
  \caption{Comparison of different types of input texts.}
  \centering
  \scalebox{0.82}{
  % @{}: delete indent
  \begin{tabular}{ccccccc}
    \toprule
    \multicolumn{3}{c}{Input text type} & \multicolumn{2}{c}{MARS} & \multicolumn{2}{c}{iLIDS-VID} \\
    Caption & \textit{ID text} & \textit{ID emb} & mAP & Rank-1 & Rank-1 & Rank-5 \\
    \midrule
    \checkmark & $\times$ & $\times$ & 89.7 & 92.3 & 96.0 & 99.9 \\
    $\times$ & \checkmark & $\times$ & 89.5 & 92.3 & 94.7 & 99.9 \\
    $\times$ & $\times$ & \checkmark & 89.6 & 92.3 & 95.3 & 99.3 \\
    \checkmark & \checkmark & $\times$ & 89.8 & 92.5 & 96.7 & 99.9 \\
    \checkmark & $\times$ & \checkmark & 89.8 & 92.4 & 96.0 & 99.9 \\
    \bottomrule
  \end{tabular}
  }
  \label{tab:ablation_text}
  % \end{minipage}
\end{table}

\begin{table}[tbp]
  % \begin{minipage}[c]{0.35\hsize}
  \caption{Effect of different caption sources.}
  \centering
  % @{}: delete indent
  \scalebox{0.87}{
  \begin{tabular}{ccccc}
    \toprule
    \multirow{2}{*}{Method} & \multicolumn{2}{c}{SportsVReID} & \multicolumn{2}{c}{DanceVReID} \\
      & mAP & Rank-1 & mAP & Rank-1 \\
    \midrule
    TF-CLIP~\cite{yu2024tf} (w/o caption) & 77.3 & 89.7 & 51.7 & 70.8 \\
    Ours: MLLM only & 77.5 & 89.3 & 53.5 & 74.2 \\
    Ours: Manual + MLLM & 77.7 & 90.4 & 53.8 & 76.0 \\
    \bottomrule
  \end{tabular}
  }
  \label{tab:ablation_caption_type}
  % \end{minipage}
\end{table}

\subsection{Effect of caption sources}
In this work, the captions used for training on SportsVReID and DanceVReID are generated based on manually annotated data provided during dataset creation.
Specifically, for SportsVReID, accurate annotations including shoe color, sock color, and jersey number are assigned to each player, ensuring high caption quality.
However, when applying our method to other datasets, obtaining such precise text annotations can be challenging.
In such cases, generating pseudo-captions using MLLMs for image captioning, as we apply to existing video-based person ReID datasets, becomes a practical solution.
Therefore, we also conduct training using captions generated solely by MLLMs for both SportsVReID and DanceVReID.
In this setting, we also append \textit{ID text} to each pseudo-caption to make the text features identity-discriminative.

As shown in \cref{tab:ablation_caption_type}, we compare the performance of different caption sources.
The results indicate that our method, which combines manual annotations with MLLM-generated captions, achieves the best performance on both datasets.
Notably, even when using only MLLM-generated captions, the performance remains competitive, demonstrating the feasibility of this approach for datasets where manual annotation is impractical.

\section{Visualization}
\subsection{Visualization of inference results}

\begin{figure}[tbp]
    \centering
    \includegraphics[width=1.0\linewidth]{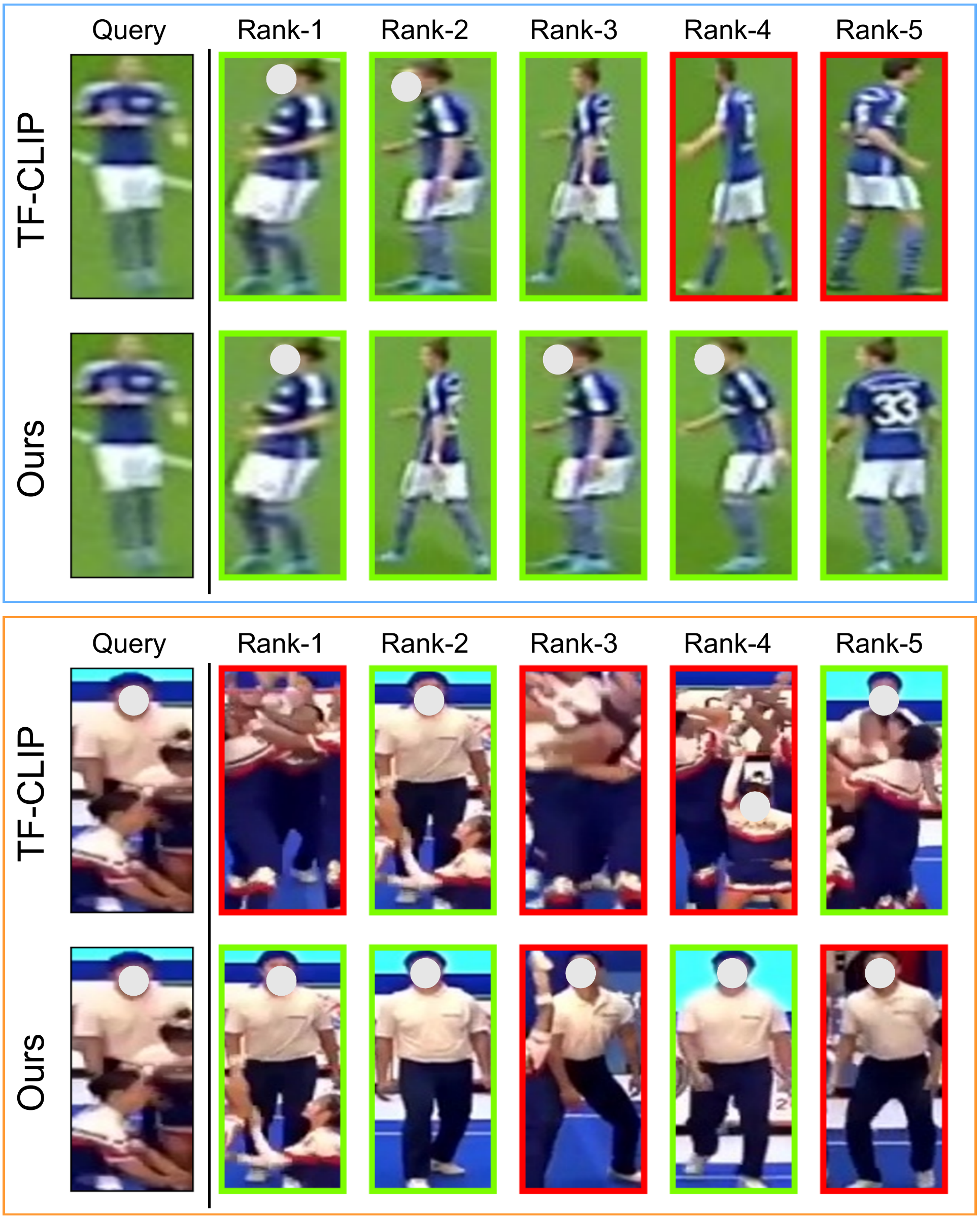}
    \caption{
      Visualization of top-5 retrieval results.
      Green and red boxes represent correct and incorrect matches, respectively.
      Blue-outlined rows show SportsVReID results, and orange-outlined rows show DanceVReID results. 
      Each image shows the first frame of a tracklet.
    }
    \label{fig:visualize_infer}
\end{figure}

To comprehensively analyze the effectiveness of our method compared to the comparative method~\cite{yu2024tf}, we visualize the ReID inference results with top-5 rankings.
As shown in \cref{fig:visualize_infer}, the top two rows with blue outlines present inference results on SportsVReID, while the bottom two rows with orange outlines show examples from DanceVReID.
Green and red boxes indicate correct and incorrect matches, respectively.
Note that only the first frame of each 8-frame tracklet is displayed for clarity.
These visualizations demonstrate our method's superior ability to handle high-difficulty scenarios where many individuals with similar appearances are present.

\subsection{t-SNE visualization of feature distributions}
\begin{figure}[tbp]
    \centering
    \includegraphics[width=1.0\linewidth]{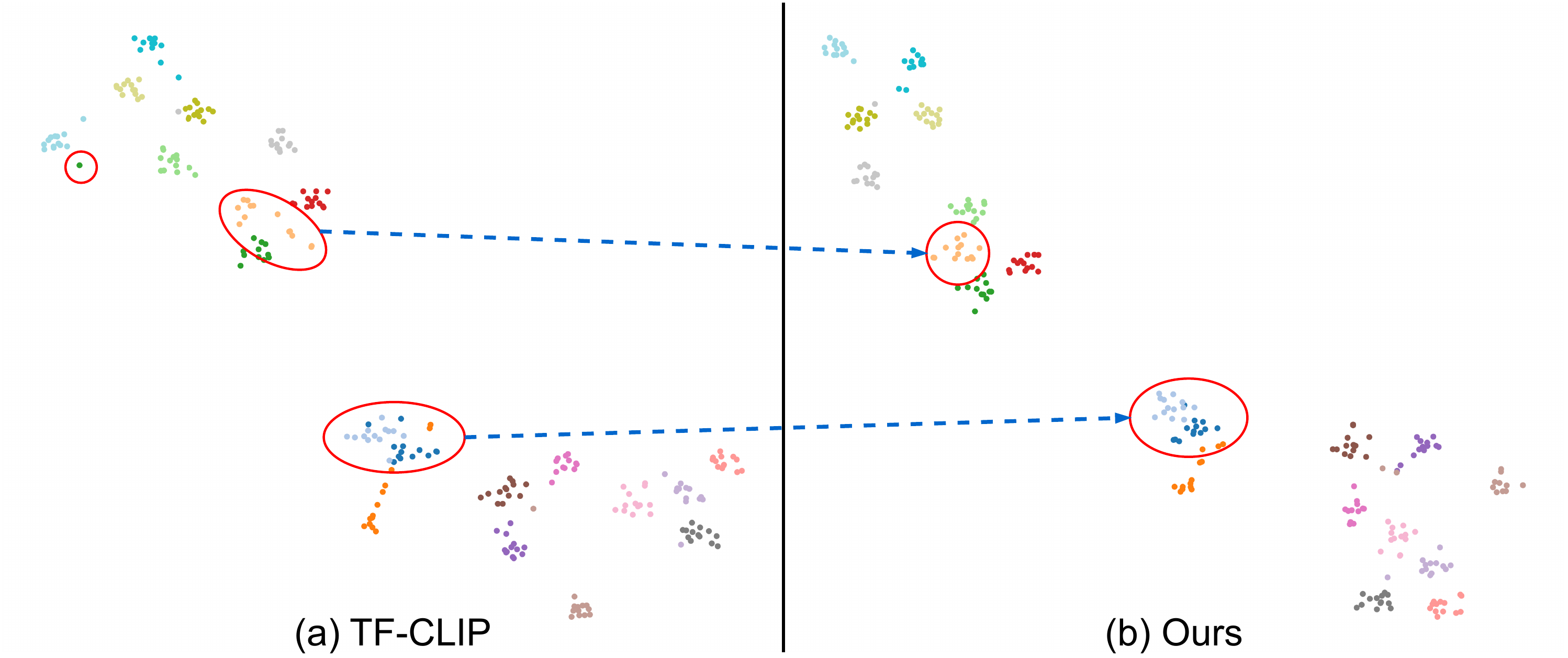}
    \caption{
      t-SNE visualization of TF-CLIP and our CG-CLIP on the DanceVReID val set.
      Different colored dots represent different identities.
      Best viewed in color.
    }
    \label{fig:visualize_tsne}
\end{figure}

To further validate the discriminative capability of our learned features, we perform t-SNE~\cite{maaten2008tsne} visualization on DanceVReID.
We sample 20 identities from three dance groups with high visual similarity and visualize the feature distributions extracted by TF-CLIP~\cite{yu2024tf} and our method in \cref{fig:visualize_tsne}.
As shown in \cref{fig:visualize_tsne} (a) and (b), our method produces more compact and well-separated clusters for several identities.
The red circles highlight specific examples where our approach achieves tighter intra-class clustering, demonstrating that our method effectively captures identity-specific features even in challenging scenarios with highly similar appearances.

\subsection{Visualization of attention in TFE}
\begin{figure}[tbp]
    \centering
    \includegraphics[width=1.0\linewidth]{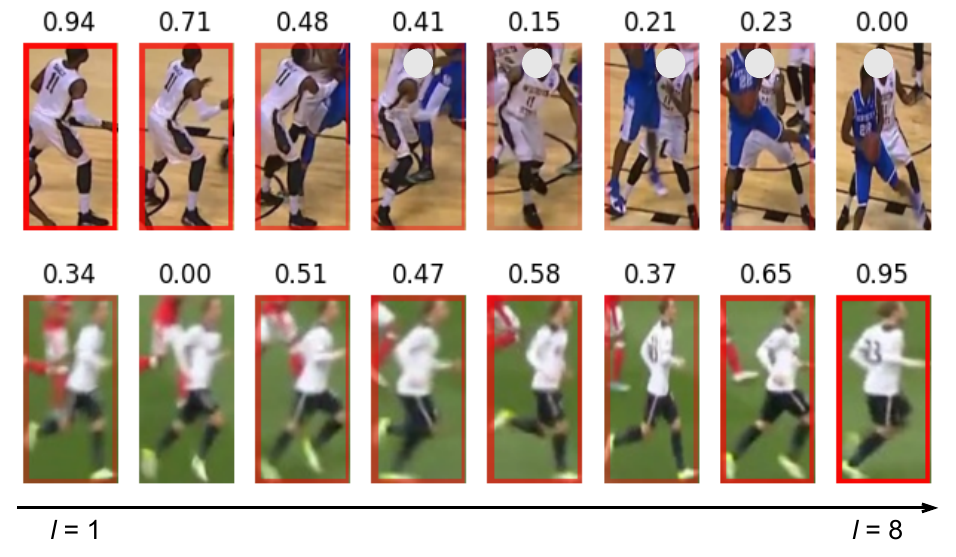}
    \caption{
      Visualization of attention weights between the first learnable token and image features from each frame in TFE.
      Higher values indicate stronger attention to the corresponding frame.
    }
    \label{fig:visualize_attention}
\end{figure}

We visualize the attention weights between the learnable tokens in the Token-based Feature Extraction (TFE) module and the image features from each input frame.
\Cref{fig:visualize_attention} presents these results, where the numerical values above each frame represent the attention weight associated with the first learnable token.
The visualization reveals that frames with severe occlusion by other people or blur due to rapid motion receive lower attention weights, while frames where the target person is clearly visible receive higher weights.
This indicates that our method effectively selects informative frames for feature aggregation.

\section{Implementation details}
We provide comprehensive training settings for all datasets in \cref{tab:training_settings}. 
Our model is implemented using PyTorch and trained on a single NVIDIA A5000 GPU with 24GB memory.
All experiments use ViT-B/16 as the image encoder from the pre-trained CLIP model.
For data augmentation, we apply random flipping and random erasing~\cite{zhong2020randomerase} during training.
The learning rate is warmed up linearly from $5 \times 10^{-7}$ to $5 \times 10^{-6}$ over the first 10 epochs.

\begin{table}[tbp]
  % \begin{minipage}[c]{0.35\hsize}
  \caption{Training settings. ``Others'' include SportsVReID and DanceVReID.}
  \centering
  % @{}: delete indent
  \scalebox{0.85}{
  \begin{tabular}{lccc}
    \toprule
    \multirow{2}{*}{Hyperparameter} & \multicolumn{3}{c}{Settings} \\
      & MARS & iLIDS-VID & Others \\
    \midrule
    Batch size & 32 & 32 & 32 \\
    \# Identities per batch & 8 & 8 & 8 \\
    \# Tracklets per identity & 4 & 4 & 4 \\
    \# Frames per tracklet & 8 & 8 & 8 \\
    \midrule
    Patch size & 16 & 16 & 16 \\
    Image size ([$H, W$]) & [256, 128] & [256, 128] & [256, 128] \\
    Max. \# text tokens & 77 & 77 & 77 \\
    \midrule
    $N^Q$ (learnable tokens) & 50 & 15 & 15 \\
    Momentum factor & 0.2 & 0.2 & 0.2 \\
    $L_{v2rm}$ weight & 1.0 & 1.0 & 1.0 \\
    $L_{tri}$ weight & 1.0 & 1.0 & 1.0 \\
    $L_{ce}$ weight & 0.25 & 0.25 & 0.25 \\
    \midrule
    Epochs & 80 & 60 & 60 \\
    Optimizer & Adam & Adam & Adam \\
    Learning rate & $5 \times 10^{-6}$ & $5 \times 10^{-6}$ & $5 \times 10^{-6}$ \\
    Weight decay & $1 \times 10^{-4}$ & $1 \times 10^{-4}$ & $2.5 \times 10^{-4}$ \\
    LR scheduler  & \multicolumn{3}{c}{StepLR ($\times 0.1$ at epochs 30, 50, 70*)} \\
    \bottomrule
    \multicolumn{4}{r}{\footnotesize *Only for MARS dataset.} \\
  \end{tabular}
  }
  \label{tab:training_settings}
  % \end{minipage}
\end{table}

\section{Limitations}
While our CG-CLIP framework demonstrates significant improvements in video-based person ReID, we acknowledge several limitations that warrant future investigation.

First, our method's performance is inherently dependent on the quality of generated captions.
MLLM-based image captioning is susceptible to hallucinations, particularly when person images have low resolution or poor clarity, which are common in surveillance and sports video scenarios.
These inaccuracies in pseudo-captions can propagate through our CMR module and potentially degrade the final person ReID performance.
Future work should focus on developing more robust caption generation methods, quality assessment mechanisms, and filtering strategies.

Second, although our method achieves substantial performance improvements over existing approaches in high-difficulty scenarios, we observe that some challenging cases remain, particularly in dance scenes, where individuals exhibit nearly identical visual attributes, making it extremely difficult to verbalize subtle differences through textual descriptions.
In such extreme scenarios, language-based descriptions face inherent limitations in expressiveness, as fine-grained visual differences may not be easily captured through natural language.
Future approaches could address this limitation by incorporating non-linguistic cues such as facial characteristics, or by developing hybrid frameworks that adaptively balance language-guided and pure visual feature learning based on scenario difficulty.

Finally, while our TFE module improves computational efficiency through its linear complexity with respect to input length, the overall inference speed of our framework is still largely governed by the image encoder.
% Although TFE successfully reduces the computational overhead of temporal aggregation, we have not yet applied optimization techniques to the image encoder itself.
For real-time applications such as online multi-object tracking, future work could explore lightweight architectures or model compression techniques, including pruning and quantization, to accelerate the image encoder while maintaining person ReID accuracy.

% {
%     \small
%     \bibliographystyle{ieeenat_fullname}
%     \bibliography{main}
% }

\end{document}